\def\year{2022}\relax
\documentclass[letterpaper]{article} 
\usepackage{aaai22}  
\usepackage{times}  
\usepackage{helvet}  
\usepackage{courier}  
\usepackage[hyphens]{url}  
\usepackage{graphicx} 
\usepackage{natbib}  
\usepackage{caption} 
\DeclareCaptionStyle{ruled}{labelfont=normalfont,labelsep=colon,strut=off} 
\usepackage{algorithm}
\usepackage{algorithmic}
\usepackage{bm} 
\usepackage{amsmath,amssymb}
\usepackage{algorithm}
\usepackage{algorithmic}
\usepackage{graphicx}
\usepackage{tabularx}
\usepackage{adjustbox}
\usepackage{svg}
\usepackage{subfigure}
\usepackage{wrapfig}
\usepackage{caption}
\usepackage{booktabs}
\usepackage{multirow}

\newcommand{\jyp}[1]{{\color{black}#1}}
\newcommand{\ysd}[1]{{\color{black}#1}}

\newcommand{\jhp}[1]{{\color{black}#1}}

\usepackage{newfloat}
\usepackage{listings}
\floatstyle{ruled}
\newfloat{listing}{tb}{lst}{}
\floatname{listing}{Listing}
\title{Deformable Graph Convolutional Networks}

\author {
    Jinyoung Park,
    Sungdong Yoo,
    Jihwan Park,
    Hyunwoo J. Kim\thanks{is the corresponding author.}
}
\affiliations {
    Department of Computer Science and Engineering, Korea University\\
    \{lpmn678, ysd424, jseven7071, hyunwoojkim\}@korea.ac.kr
}

\usepackage{bibentry}

\newcommand{\omitme}[1]{}

\newcommand{\Lc}{\mathcal{L}}

\def\ie{\emph{i.e}.}
\def\xb{\mathbf{x}}

\def\yb{\mathbf{y}}

\def\hb{\mathbf{h}}

\def\eb{\mathbf{e}}
\def\Wb{\mathbf{W}}
\def\zb{\mathbf{z}}

\def\rb{\mathbf{r}}

\def\Rbb{\mathbb{R}}

\def\Gc{\mathcal{G}}

\def\Nc{\mathcal{N}}

\def\phib{\boldsymbol{\phi}}

\def\Nc{\mathcal{N}}

\def\phib{\bm{\phi}}

\newcommand{\Hc}{\mathcal{H}}

\newcommand{\gdeform}{g_\text{deform}}

\makeatletter
\newcount\my@repeat@count
\newcommand{\myrepeat}[2]{%
  \begingroup
  \my@repeat@count=\z@
  \@whilenum\my@repeat@count<#1\do{#2\advance\my@repeat@count\@ne}%
  \endgroup
}
\makeatother

\begin{document}

\maketitle

\begin{abstract}
    Graph neural networks (GNNs) have significantly improved the representation power for graph-structured data.
Despite of the recent success of GNNs, the graph convolution in most GNNs
have two limitations. 
Since the graph convolution is performed in a small local neighborhood on the input graph, it is inherently incapable to capture long-range dependencies between distance nodes. 
In addition, when a node has neighbors that belong to different classes, \ie, \textit{heterophily}, the aggregated messages from them often negatively affect representation learning.
To address the two common problems of graph convolution, in this paper, 
we propose Deformable Graph Convolutional Networks~(Deformable GCNs) that adaptively perform 
convolution in multiple \textit{latent spaces} and capture short/long-range dependencies between nodes.
Separated from node representations (features), our framework simultaneously learns the \textit{node positional embeddings} (coordinates) to determine the relations between nodes in an end-to-end fashion.
Depending on node position, the convolution kernels are deformed by deformation vectors and apply different transformations to its neighbor nodes.
Our extensive experiments demonstrate that Deformable GCNs flexibly handles the heterophily and achieve the best performance in node classification tasks on six  heterophilic graph datasets.

\end{abstract}
\section{Introduction}
Graphs are flexible representations for modeling relations in data analysis problems and are widely used in various domains such as social network analysis~\cite{wang2016structural}, recommender system~\cite{berg2017graph}, chemistry~\cite{gilmer2017neural}, natural language processing~\cite{erkan2004lexrank}, and computer vision~\cite{johnson2015image}.
In recent years, Graph Neural Networks~(GNNs) have achieved great success in many graph-related applications such as node classification~\cite{kipf2016semi, hamilton2018inductive}, link prediction~\cite{zhang2018link, schlichtkrull2018modeling}, and graph classification ~\cite{errica2019fair, ying2018hierarchical}.
Most existing GNNs learn node representations via message passing schemes, which iteratively learn the hidden representation of each node by aggregating messages from its local neighborhoods.
For example, Graph Convolution Networks (GCNs)~\cite{kipf2016semi} operate convolutions on input graphs inspired by first-order approximation of spectral graph convolutions~\cite{hammond2011wavelets}.

However, most graph convolution that aggregates messages from local neighborhoods
implicitly assumes that input graphs are homophilic graphs, where connected nodes have similar features or belong to the same class. 
So the smoothing over the input graphs effectively removes noise in the input features and significantly improve the representational power when the assumption holds. 
However, on heterophilic graphs where connected nodes have dissimilar features and different labels, 
the conventional graph convolutional neural networks often underperform simple methods such as a multi-layer perceptron (MLP) that completely ignores the graph structure. 
In addition, since the conventional graph convolution receives messages from local neighbors, 
it has the limited capability to capture long-range dependencies between distant yet relevant nodes for the target tasks.

To address these limitations, we propose a Deformable Graph Convolutional Network~(Deformable GCN) that softly changes the receptive field of each node by adaptively aggregating the outputs of deformable graph convolution in multiple latent spaces.
Started from a general definition of the discrete convolution with finite support, 
we extend the deformable 2D convolution ~\cite{dai2017deform} to a latent space for graph-structured data.
Similar to the convolution defined on a grid space for images, our convolution kernel generates different transformations 
for various relations. Our framework models useful relations between nodes represented by the difference of learned \textit{node positional embeddings}.
Our \textbf{contributions} are as follows:
\begin{itemize}
    \item We propose a Deformable Graph Convolution~(Deformable GConv) that performs convolution in a latent space and adaptively deforms the convolution kernels to handle heterophily and variable-range dependencies between nodes.
    \item We propose novel architecture Deformable Graph Convolution Networks (Deforamble GCN) that simultaneously 
    learn node representations (features) and node positional embeddings (coordiantes) and efficiently perform
     Deformable GConv in multiple latent spaces using latent neighborhood graphs. 
    \item Our experiments demonstrate the effectiveness of Deformable GCN in the node classification task on homophily and heterophily graphs.  
    Also, the \textit{interpretable} attention scores in our framework provide insights which relation (or latent space) is beneficial to the target task.
\end{itemize}
\section{Related Works}

\noindent\textbf{Graph Neural Networks.}
Graph neural networks have been studied for representing graph-structured data in recent years.
Based on spectral graph theory, ChebyNet~\cite{defferrard2016convolutional} designed fast localized convolutional filters and reduced computational cost.
Motivated by it, \cite{kipf2016semi} proposed GCN, which is simplified convolutional networks based on the first-order approximation of the spectral convolutions.
There are several studies to improve the performance by  message passing processes \cite{hamilton2018inductive, xu2018representation} and attention-based models \cite{velivckovic2017graph, yun2019gtn}. 

However, the studies introduced above have been developed based on the {\em homophily assumption} that connected nodes have similar characteristics. This assumption often leads to performance degradation on {\em heterophilic graphs}.
Moreover, most existing GNNs cannot capture the {\em long-range dependencies} between distant nodes because they aggregate messages only in a local neighborhood.
\ysd{Recently, there are several studies have attempted to address the problems \cite{bo2021beyond, liu2020non, zhu2020beyond, zhu2021hetero}.  
Geom-GCN \cite{pei2020geom} proposed a novel geometric aggregation scheme in a latent space that captures long-range dependencies through structural information.
However, Geom-GCN has some limitations that they define convolution on a grid that is a manually divided latent space and require pre-trained embedding methods in the geometric aggregation step. 
Unlike Geom-GCN, our models apply deformable convolution kernel on the latent space with the relation vectors defined in a continuous latent space and utilize learnable multiple embeddings to softly grant relations in an end-to-end fashion.
}

\paragraph{Deformable Convolution.}
Convolutional neural networks (CNNs) have achieved great success in various fields~\cite{he2016deep,he2017mask}.
However, the convolution kernels are limited to model large and unknown transformations since they are defined in a fixed structure. 
To address these limitations, \cite{dai2017deform, zhu2019deformable} proposed deformable convolution networks that adaptively change the shape of convolution kernels by learning offsets for deformation 
Because of the feasibility and effectiveness, deformable convolution networks are used in various fields, such as point cloud \cite{thomas2019kpconv}, image generation \cite{siar2018deformgan}, and video tasks \cite{wang2019videorestor}. 
Inspired by this line of work, we study a deformable graph convolution to adaptively handle diverse relations between an ego-node and its neighbors.
\section{Method}

%
%
%
%
%
%
%
%
The goal of our framework is to address the limitations of existing graph convolutions that learn node representations in a small neighborhood with the homophily assumption. 
In other words, existing GNNs often poorly perform when neighbors belong to different classes and have dissimilar features. 
Also, most GNNs with a small number of layers cannot model the long-range dependency between distant nodes. 
To address the limitations, we propose Deformable Graph Convolutional Networks that perform deformable convolution on latent graphs. 
Our framework softly changes the shape (or size) of the receptive field for each node.
This allows our framework to adaptively handle homophilic/heterophilic graphs as well as short/long-range dependency.
Before introducing our frameworks, we first briefly summarize notations and the basic concepts of graph neural networks.

\subsection{Preliminaries}
\paragraph{Notations.}
Let $G = (V, E)$ denote an input graph with a set of nodes $V$ and a set of edges $E \subseteq V \times V$.
Each node $v \in V$ has a feature vector $\xb_v \in \Rbb^{d_x}$ and the edge between node $u$ and node $v$ is represented by $(u, v) \in E$.
The neighborhoods of node $v$ on input graph is denoted by $N\left(v\right) = \left\{u\in V|\left(u, v\right) \in E\right\}$.
\begin{figure*}[ht]
\centering
\includegraphics[width=1\textwidth]{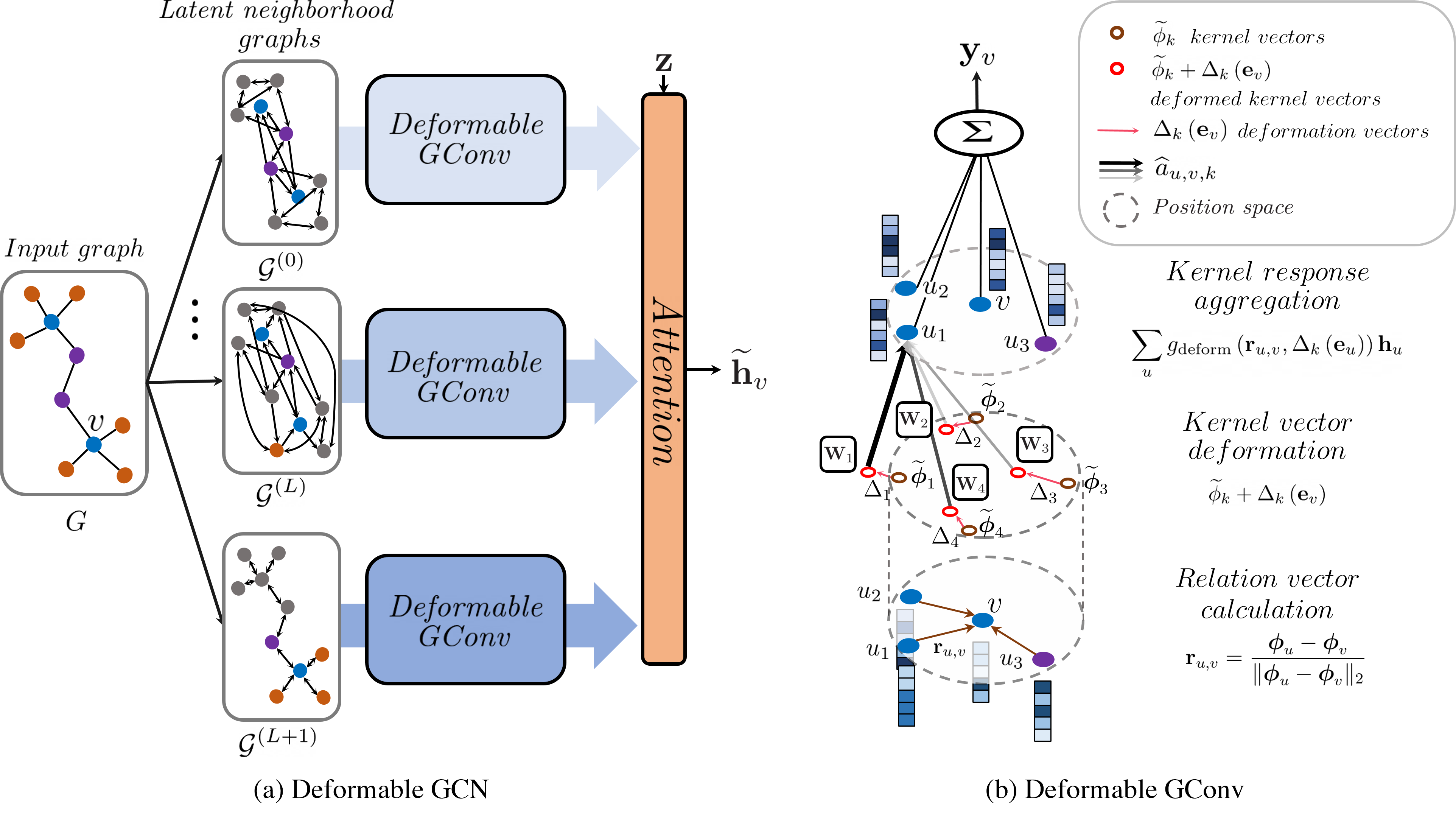}
\caption{\jhp{Overall structure of Deformable GCN and Deformable GConv.
(a) In Deformable GCN, at the beginning of training, latent neighborhood graphs, $\{\Gc^{(l)}\}_{l=0}^{L+1},$ are constructed to define neighborhoods for the Deformable Graph Convolution~(Deformable GConv). 
Then, Deformable GConv is applied on each latent neighborhood graph and the outputs of the convolution $\{\yb_v^{(l)}\}_l$ are adaptively aggregated for representing $\widetilde{\mathbf{h}}_v$ using an attention mechanism.
\textcolor{black}{(b) Our Deformable GConv performs graph convolution in a latent (position) space. For more flexible graph convolution, Deformable GConv adaptively deforms convolution kernels $g_{\text{deform}}(\cdot,\cdot)$ for each center node $v$ by kernel vector deformation 
$\Delta_k(\mathbf{e}_v)$. 
}
}
}
\label{fig:temp_overall}
\end{figure*}

\paragraph{Graph Neural Networks.} 
To learn representation for graph-structured data, 
most existing GNNs perform message-passing frameworks as the following equation~\cite{gilmer2017neural,xu2018representation}: 
\begin{equation}
\label{eq:mpnn}
\begin{split} 
        \mathbf{h}_v^{(l)} = \sigma \left(\Wb^{(l)}\cdot \text{AGGREGATE} \left(\mathbf{h}_u^{(l-1)}: u \in \widetilde{N}\left(v \right) \right) \right), 
\end{split}
\end{equation}
where $\mathbf{h}_v^{(0)} = \mathbf{x}_v$, $\mathbf{h}_v^{(l)} \in \Rbb^{d_{h^{(l)}}}$ is a hidden representation of node $v$ in a $l$-th layer, $\widetilde{N}\left( v \right) = \{v\}\cup\{u \in V | (u,v)\in E\}$ indicates neighbors of node $v$ with a self-loop, $\Wb^{(l)}$ is a learnable weight matrix at the $l$-th layer, AGGREGATE is an aggregation function characterized by the particular model, and $\sigma$ is a non-linear activation function. 
$\widetilde{N}\left( v \right)$ determines the receptive field of the graph convolution and it is usually a one-hop ego-graph.
For example, GCN~\cite{kipf2016semi} is a specific instance of \eqref{eq:mpnn}. GCN can be written as 
\begin{equation}
\label{eq:gcn}
        \mathbf{h}_v^{(l)} = \sigma  \left ( \sum_{u \in \widetilde{N}(v)} (\deg(v)\deg(u))^{-1/2}\ \Wb^{(l)} \hb_u^{(l-1)} \right ),
\end{equation}
where $\deg(v)$ is the degree of node $v$.

Even GNNs such as GCN~\cite{kipf2016semi} and GAT~\cite{velivckovic2017graph} have been successfully applied to various graph-based tasks, their success is limited to \textit{homophilic} graphs~\cite{pei2020geom,zhu2020beyond}, which are graphs that linked nodes often have similar properties~\cite{mcpherson2001birds}.
Recent works~\cite{li2018deeper,wu2019simplifying} showed that GCN makes node embeddings smoother within their peripherals since GCN can be a specific form of Laplacian smoothing~\cite{li2018deeper}.
For this reason, graph neural networks have difficulty adapting to graphs that linked nodes often have different properties, called \textit{heterophilic} graphs.
 
In our work, we consider both heterophilic and homophilic graphs, unlike standard graph neural networks that have mainly focused on homophilic graphs.
To have enough representation power on heterophilic graphs, we generate latent graphs for linking distant nodes with similar property according to their latent embeddings.

\subsection{Deformable Graph Convolution}


We here introduce a Deformable Graph Convolution~(Deformable GConv), which softly changes the receptive fields and adaptively aggregates messages from neighbors on the multiple latent graphs. 
The  overall structure of Deformable GConv is illustrated in Figure~\ref{fig:temp_overall}.
Our Deformable Graph Convolution can be derived from a general definition of the discrete convolution by a kernel with finite support. 
The convolution of a feature map $\Hc$ by a kernel $g$ at a point $v$ is given as:
\begin{equation}
\label{eq:gen_conv}
    \yb_v = \left(\Hc*g \right)\left(v\right) = \sum_{u \in \widetilde{\Nc}\left(v\right)} g\left(\rb_{u,v} \right)\hb_u,
\end{equation}
where $\yb_v \in \Rbb^{d_y}$ is the output of the convolution, $\hb_u \in \Rbb^d$ is a feature at $u$, and relation vector $\rb_{u,v}$ represent the relation between $u$ and $v$, $\widetilde{\Nc}(v)$ is the neighborhood of $v$ that coincides with the finite support of $g$ centered at $v$. 
$g(\rb_{u,v})$ is a linear function to transform $\hb_u$ and it varies depending on the relation vector.
For example, in a 2D convolution with a $3 \times 3$ kernel,  $\rb_{u,v} = \phib_u - \phib_v$, where $\phib_u$, $\phib_v$ are the coordinates of $u$ and $v$. For each relative position, a linear function  $g(\rb_{u,v})\in \Rbb^{d_y \times d_h}$ is applied.

In the graph domain, a GCN layer defined in \eqref{eq:gcn} (without the activation function) 
can be viewed as a specific instantiation of \eqref{eq:gen_conv} with $g(\rb_{u,v})= (\deg(v)\deg(u))^{-1/2}\Wb$. 
So, the GCN relations are determined by the degree of node $u$ and $v$. 
Also, except for the normalization, GCN performs the same linear transformation for all the relations different from standard 2D convolution. 

Our framework extends the graph convolution to more flexible and deformable graph convolutions defining a relation vector $\rb_{u,v}$, a kernel function $g(\cdot)$, and its support (or a neighborhood). 
To extend the relation of nodes beyond the adjacency on the input graph $G$, 
we first embed nodes in a position space, which is a latent space to determine the coordinates of nodes using a \textit{node positional embedding}.
Then, we compute the \textit{relation vector} of the nodes by a function of the node positional embeddings.

\paragraph{Node positional embedding.}
In our framework, each node is embedded in a latent space called \textit{position space} using a node embedding method. 
Since the node embeddings are used only for the coordinates of nodes and their relations, we name this \textit{node positional embedding}. 
For each node, our framework learns both node positional embeddings (coordinates) and node representations (features).

Node positional embedding is computed by a simple procedure. 
Given a node $v$, its node positional embedding $\phib_v^{(l)}$ is the projected input features after smoothing $l$ times on the original input graph $G$. 
It can be written as 
\begin{equation}
\label{eq:position}
\begin{split}
    \phib_v^{(l)} &= \Wb_\phi^{(l)} \eb_v^{(l)} \text{, where }\\
    \eb_{v}^{(0)} &= \mathbf{x}_{v}, \quad \eb_{v}^{(l)} = \frac{1}{\widetilde{\text{deg}}(v)}\sum_{u \in {\widetilde{N}}(v)}\eb_{u}^{(l-1)},
\end{split}
\end{equation}
where $\Wb_\phi^{(l)}$ is a learnable project matrix, $\widetilde{\text{deg}}\left(v \right)$ indicates a degree of node $v$ with self-loop and $\eb_v^l$ is the $l$-time smoothed input features of node $v$.
Each node has $L+1$ node positional embeddings. 
Alternatively, for node positional embedding $\phib_v$, other node embedding methods can be used such as LINE~\cite{tang2015line}, Node2Vec~\cite{grover2016node2vec}, distance encoding~\cite{li2020distance}, Poincare embedding in hyperbolic geometry \cite{nickel2017poincare}, and other various node embedding methods.
But our preliminary experiment showed that our simple node positional embedding was sufficient for our model. 
So in this paper, we did not use any external node embedding method.

\paragraph{Relation vector.} The relation between nodes is represented by a relation vector $\rb_{u,v}$. 
One natural choice is the relative position of neighbor node $u$ from node $v$ in the position space. 
In our framework, we use normalized relation vectors with an extra dimension to encode the relation of nodes with identical positional embeddings.
The relation vector for a neighbor node $u$ of node $v$ with node positional vectors $\phib_u, \phib_v \in \Rbb^{d_\phi}$ is given as 
\begin{equation}
\label{eq:relvec}
    \rb_{u,v} = 
    \begin{cases}
        \left[\rb_{u,v}^{\prime}||0\right] \in \Rbb^{d_{\phi+1}}, &\text{if } \phib_u \ne \phib_v \\
        \left[0,0,\cdots, 1\right] \in \Rbb^{d_{\phi+1}}, &\text{ otherwise}
    \end{cases}
\end{equation}
where $\rb_{u,v}^\prime = \frac{\phib_{u}-\phib_{v}}{\lVert \phib_{u}-\phib_{v}\rVert_2}$ and $\left[||\right]$ is concatenation operator.

\paragraph{Kernel function.} As discussed with \eqref{eq:gen_conv}, a kernel function $g$ yields linear functions to transform hidden representations of neighbor nodes. 
Our kernel function $g$ on $\rb_{u,v}$ is defined as:
\begin{equation}
\label{eq:ours_conv}
\begin{split}
    &g\left(\rb_{u,v} \right) = \sum_{k=1}^K a_{u,v,k} \Wb_k,\\ &\text{ where }a_{u,v,k} = \text{exp}\left(\rb_{u,v}^\top \widetilde{\phib}_k \right)/Z.
\end{split}
\end{equation}
$\widetilde{\phib}_k \in \Rbb^{d_{\phi}+1}$ is a kernel vector and $\Wb_k \in \Rbb^{d_y \times d_h}$ is  a corresponding transformation matrix.  Both are learnable parameters. $Z \in \Rbb$ is a normalization factor.
The function value of $g$, which is a linear transformation, varies depending on the relation vector $\rb_{u,v}$ but for the same relation, the same linear transformation is returned when $Z$ is a constant. 
This is the same as a standard 2D convolution that applies the identical linear transformation at the same relative position from the center of the kernel.
One interesting difference 
is that since a standard 2D convolution kernel has a linear transformation matrix for each relative position, the number of its transformation matrices increases as the kernel size increases 
whereas our kernel $g$ differentially combines a fixed number of $\{\Wb_k\}_{k=1}^K$ matrices depending on the relation vector $\rb_{u,v}$.
Thereby the number of parameters of our convolution does not depends on the kernel size anymore.

\paragraph{Deformable Graph Convolution.} The kernel function can be further extended for a more flexible and deformable convolution.
We first normalize the weight $a_{u,v,k}$ for a transformation matrix $\Wb_k$ by $Z = \sum_{u'}\text{exp}\left(\rb_{u',v}^\top \widetilde{\phib}_k \right)$. Then the kernel becomes adaptive. 
This is similar to the dot-product attention in  \cite{vaswani2017attention}.
We found that the dot-product attention can be viewed as a variant of convolution. For more details, see the supplement. 
Now the kernel yields a transformation matrix for each neighbor considering not only the relation between a center node $v$ and a neighbor node $u$ but also the relations between neighbor nodes. 
In addition, the kernel point $\widetilde{\phib}_k$ is dynamically translated by deformation vector $\Delta_k(\eb_v) \in \Rbb^{d_{\phi}+1}$ depending on the smoothed input features $\eb_v\in \Rbb^{d_x}$ of the center node $v$.
Putting these pieces together, we define our Deformable Graph Convolution as:
\begin{equation}
\yb_v =\sum_{u \in \widetilde{\Nc}\left(v\right)} g_{\text{deform}}\left(\rb_{u,v}, \Delta_k\left({\eb_v}\right) \right)\hb_u, 
\end{equation}
where $\gdeform \left(\rb_{u,v}, \Delta_k\left(\eb_v\right) \right) = \sum_{k=1}^K \widehat{a}_{u,v,k} \Wb_k$ and $ \widehat{a}_{u,v,k} = \frac{\text{exp}\left(\rb_{u,v}^\top \left(\widetilde{\phib}_k + \Delta_k\left(\eb_v \right) \right)\right)}{\sum_{u'}\text{exp}\left(\rb_{u',v}^\top \left ( \widetilde{\phib}_k + \Delta_k\left(\eb_v \right) \right)\right)}$.
In our experiments, the deformation vector is generated by a simple MLP network with one hidden layer.


%
\paragraph{Latent Neighborhood Graph.}
The Deformable Graph Convolution is computed within a neighborhood $\widetilde{\Nc}(v)$.
To efficiently compute the neighborhoods, $k$-nearest neighbors for each node are computed with respect to $\ell_2$ distance of smoothed input features, \ie, $\|\eb_u- \eb_v\|_2$.
Since in our framework the input feature smoothing does not have any learnable parameters, the smoothed features $\{\eb_u \}$ do not change during training. 
Thereby, the kNN graph generation is performed once at the beginning of training.
Due to the kNN graphs, our Deformable Graph Convolution can be viewed as a graph convolution on the \textcolor{black}{kNN} graphs.
Seemingly, the neighborhood computed by node positional embedding, $\phib_v$, is more plausible following \cite{dai2017deform}.
But, it requires huge computational costs for pairwise distance computation of all nodes at every iteration. 
Also, in practice, the drastic changes of neighborhoods caused by positional embedding learning often lead to unstable numerical optimization.
So, in this work, we use the smoothed input features $\left\{\eb_v \right\}_{v}$ for the neighborhood computation.


\subsection{Deformable Graph Convolutional Networks}

With our Deformable Graph Convolution, we design our Deformable Graph Convolution Network (Deformable GCN).
The overall structure is depicted in Figure~\ref{fig:temp_overall}.
Deformable GCN utilizes multiple node positional embeddings. In other words, it generates multiple neighborhood graphs.
Let $\eb_v^{(l)}$ denote the input features at node $v$ that are smoothed $l$ times on the original graph $G$. 
A Latent Neighborhood Graph, \ie, kNN graph, constructed based on $\{\eb_v^{(l)}\}_{v\in V}$ is denoted by $\Gc^{(l)}$.
Deformable GCN generates $L$+1 kNN graphs $\{\Gc^{(l)}\}_{l=0}^{l=L}$.
Combining with the original input graph $\Gc^{(L+1)}=G$, Deformable GCN performs the Derformable GConv on each neighborhood graph. 
The outputs of convolution on $\Gc^{(l)}$, denoted by $\yb^{(l)}_v$, are adaptively aggregated on each node by a simple attention mechanism as:
\begin{equation}
\label{eq:score}
    \widetilde{\hb}_v = \sum_{l=0}^{L+1} s_v^{(l)} \widetilde{\yb}_v^{(l)}, \quad
    s_{v}^{(l)} = \frac{\exp \left(\zb^\top \widetilde{\yb}_v^{(l)} \right)}{\sum_{l^\prime=0}^{L+1}\exp \left( \zb^\top \widetilde{\yb}_v^{(l^\prime)} \right)},
\end{equation}
where $\widetilde{\yb}_v^{(l)} = \frac{\yb_v^{(l)}}{\lVert \yb_v^{(l)} \rVert_2}$ and $\zb \in \Rbb^{d_y}$ is a learning parameter.
The score $s_v^{(l)}$ indicates which neighborhood (with its Deformable GConv) is suitable for node $v$.
For each node, Deformable GCN softly chooses a suitable neighborhood with node positional embeddings, and Deformable GConv performs convolution with a deformed convolution kernel. 

\paragraph{Loss functions.} 
To learn more effective deformable graph convolution in a latent space without collapsed kernel points, we impose two regularizers: a \textit{separating} regularizer and a \textit{focusing} regularizer. 
The \textit{separating} regularization loss maximizes the distance between kernel vectors $\{ \widetilde{\phib}_k \}_{k=1}^{k=K}$ so that our kernel differentially yields transformation matrices against diverse relations. It is formulated as
\begin{equation}
\label{eq:sep}
    \Lc_{sep.} = -\frac{1}{K}\sum_{k_1\ne k_2}\left\lVert \widetilde{\phib}_{k_2} - \widetilde{\phib}_{k_1}\right\rVert_2^2.
\end{equation}
In addition, to avoid extreme changes of the deformable kernel $\gdeform$ and the collapse of kernel points after deformations, we use  a \textit{focusing} regularizer that penalizes the $\ell_2$-norm of deformation vectors $\Delta_k \left(\eb_v \right)$ as 
\begin{equation}
\label{eq:foc}
    \Lc_{focus} = \frac{1}{K \cdot \lvert V\rvert} \sum_{v \in V}\sum_{k=1}^{K} \left\lVert \Delta_{k}(\eb_{v})\right\rVert_2^2.
\end{equation}
With these two regularizer losses, our Graph Deformable Convolutional Network properly generates the kernel vectors and deformation vectors. 
Since we mainly conduct experiments on node classification tasks, we use the standard cross-entropy loss function $\Lc_{cls}$ with the two regularizers. 
The total loss function is as follows:
\begin{equation}
    \Lc = \Lc_{cls} + \alpha\cdot\Lc_{sep.} + \beta\cdot\Lc_{focus},
\end{equation}
where $\alpha$ and  $\beta$ are hyperparameters for the strength of regularizations, $\Lc_{sep.}$ and $\Lc_{focus}$.

\section{Experiments}
\begin{table*}[ht]
  \centering
  \setlength{\tabcolsep}{4pt}
  \begin{tabular}{c|cccccc|ccc}
    \toprule
          & \multicolumn{6}{c|}{Heterophilic Graphs} & \multicolumn{3}{c}{Homophilic Graphs} \\
    Dataset    & Texas & Wisconsin  & Actor & Squirrel & Chameleon & Cornell & Citeseer & Pubmed & Cora \\
    \midrule
    \midrule
    \# Classes     &      5 &       5 &       5 &       5 &      5 &         5 &        6 &      3 & 7\\
    \# Nodes       &    183 &     251 &    7600 &    5201 & 2277 & 183 & 3327 &19717& 2708 \\ 
    \# Edges       &    279 &     450 &   26659 &  198353 & 31371 & 277 & 4552 &44324& 5278 \\
    \# Features    &   1703 &    1703 &     932 &    2089 & 2325 & 1703 & 3703 &500& 1433 \\
    Avg deg.       &   3.05 &    3.59 &    7.02 &   76.28 & 27.55 & 3.03  & 3.03 &4.50  & 3.90 \\ 
    Hom. ratio $h$ &   0.11 &    0.21 &    0.22 &    0.22 &  0.23 &  0.30 & 0.74   & 0.80 & 0.81 \\
    \bottomrule
  \end{tabular}
  \caption{Dataset statistics.}
  \label{tab:data_statistics}
\end{table*}

\begin{table*}[t]
    \centering
    \setlength{\tabcolsep}{2.5pt}    
    \begin{tabular}{l|cccccc|ccc}
    \toprule
    Dataset   & Texas & Wisconsin & Actor & Squirrel & Chameleon & Cornell & Citeseer & Pubmed & Cora \\
    Hom. ratio $h$       &  0.11 &      0.21 &  0.22 &     0.22 &    0.23 &    0.30 & 0.74 &  0.80 & 0.81 \\
    \midrule
    \midrule
    MLP       & 82.16{\scriptsize$\pm$2.64} &    84.90{\scriptsize$\pm$1.82} &  36.78{\scriptsize$\pm$0.56} &     30.77{\scriptsize$\pm$1.68} &  47.87{\scriptsize$\pm$1.31} &   81.08{\scriptsize$\pm$3.62} & 72.86{\scriptsize$\pm$1.42} &87.62{\scriptsize$\pm$0.19} & 75.09{\scriptsize$\pm$1.45}\\
    GCN       & 64.32{\scriptsize$\pm$2.60} &    62.94{\scriptsize$\pm$4.19} &  30.47{\scriptsize$\pm$0.64} &     46.65{\scriptsize$\pm$0.99} &  64.98{\scriptsize$\pm$0.69} &   60.27{\scriptsize$\pm$2.37} & 76.66{\scriptsize$\pm$1.12} &87.59{\scriptsize$\pm$0.36} & 87.44{\scriptsize$\pm$0.83}\\
    GAT       & 60.81{\scriptsize$\pm$4.62} &    64.31{\scriptsize$\pm$3.48} &  29.92{\scriptsize$\pm$0.43} &     45.47{\scriptsize$\pm$1.38} &  66.56{\scriptsize$\pm$0.99} &   59.46{\scriptsize$\pm$1.58} & 76.54{\scriptsize$\pm$1.00} &86.55{\scriptsize$\pm$0.34} & 87.46{\scriptsize$\pm$0.77}\\
    ChebyNet & 76.49{\scriptsize$\pm$3.45} &    77.84{\scriptsize$\pm$3.29} &  35.03{\scriptsize$\pm$0.73} &     45.42{\scriptsize$\pm$0.06} &  61.80{\scriptsize$\pm$1.19} &   72.97{\scriptsize$\pm$4.61} & 76.20{\scriptsize$\pm$1.12} &89.16{\scriptsize$\pm$0.30} & 83.44{\scriptsize$\pm$2.04}\\    
    JKNet     & 62.97{\scriptsize$\pm$5.18} &    60.78{\scriptsize$\pm$3.29} &  30.78{\scriptsize$\pm$0.60} &     53.78{\scriptsize$\pm$1.25} &  68.53{\scriptsize$\pm$1.59} &   59.19{\scriptsize$\pm$2.16} & 76.05{\scriptsize$\pm$1.00} &88.64{\scriptsize$\pm$0.34} & 87.12{\scriptsize$\pm$0.81}\\
    MixHop     & 83.78{\scriptsize$\pm$3.35} &    85.10{\scriptsize$\pm$2.57} &  33.80{\scriptsize$\pm$0.89} &     39.32{\scriptsize$\pm$2.16} &  63.09{\scriptsize$\pm$1.07} &  81.08{\scriptsize$\pm$4.61}  & 75.86{\scriptsize$\pm$1.20} &89.02{\scriptsize$\pm$0.26} & 87.20{\scriptsize$\pm$0.85}\\
    Geom-GCN  & 68.11{\scriptsize$\pm$3.04} &    64.51{\scriptsize$\pm$2.53} &  31.48{\scriptsize$\pm$0.64} &     38.00{\scriptsize$\pm$0.60} &  60.99{\scriptsize$\pm$1.74} &   60.54{\scriptsize$\pm$3.55} & \textbf{77.48{\scriptsize$\pm$0.87}} &\textbf{89.51{\scriptsize$\pm$0.33}} & 85.51{\scriptsize$\pm$0.92}\\    
    $\text{H}_2$GCN     & 82.16{\scriptsize$\pm$4.12} &    86.67{\scriptsize$\pm$2.18} &  36.96{\scriptsize$\pm$0.55} &     54.51{\scriptsize$\pm$0.94} &  65.42{\scriptsize$\pm$1.58} &   80.54{\scriptsize$\pm$3.77} & 77.05{\scriptsize$\pm$0.9} &89.38{\scriptsize$\pm$0.26} & \textbf{87.48{\scriptsize$\pm$0.93}}\\
    \midrule
    Deformable GCN & \textbf{84.32{\scriptsize$\pm$3.42}} &  \textbf{87.06{\scriptsize$\pm$2.16}} &  \textbf{37.07{\scriptsize$\pm$0.79}} &  \textbf{62.56{\scriptsize$\pm$1.31}} &  \textbf{70.90{\scriptsize$\pm$1.12}}  &  \textbf{85.95{\scriptsize$\pm$2.71}} &76.83{\scriptsize$\pm$1.15} & 89.49{\scriptsize$\pm$0.29} & \textbf{87.48{\scriptsize$\pm$0.82}}\\
    \bottomrule
  \end{tabular}
  \caption{
  Evaluation results on node classification task~(Mean accuracy~(\%) $\pm$ 95\% confidence interval).
  The best-performing models are highlighted with boldface.}
  \label{tab:main table}
\end{table*}

In this section, we validate the effectiveness of our framework, Deformable GCN, using heterophily and homophily graph datasets.
\subsection{Dataset}
\label{sec:datasets}
For validating our model, we use six heterophilic graph datasets and three homophilic graph datasets, which can be distinguished by the \textit{homophily ratio}~\cite{zhu2020beyond} $h = \frac{\lvert \left\{(u,v):(u,v)\in E \land y_u = y_v \right\}\rvert}{\lvert E\rvert}$, where $y_v$ is the label of node $v$.
Statistics of each dataset are in Table \ref{tab:data_statistics}.

\paragraph{Heterophilic graphs.}
We utilize six datasets for evaluating performance on \textit{heterophilic graphs}, which have low homophily ratio.
Texas, Wisconsin, and Cornell are web page datasets~\cite{pei2020geom} collected by CMU. 
Nodes correspond to web pages and edges correspond to hyperlinks between them.
Node features are the bag-of-words representations of the web pages and labels are five categories.

Squirrel and Chameleon are web page datasets in Wikipedia~\cite{rozemberczki2021multi}, where the nodes are web pages, edges are links between them, node features are keywords of the pages, and labels are five categories based on the monthly traffic of the web pages.

Actor dataset is a subgraph of the film-director-actor-writer network \cite{tang2009social}.
Each node represents an actor, edges represent co-occurrence on the same Wikipedia page, and node features are keywords on the Wikipedia page.
The number of node labels is five.

\paragraph{Homophilic graphs.}
We utilize three standard citation graphs such as Citeseer, Pubmed, and Cora~\cite{sen2008collective} for evaluation on \textit{homophilic graphs}.
In these graph datasets, nodes represent documents and edges represent citations. 
Node features are the bag-of-words representations of papers and node labels are the topics of each paper.



\subsection{Baselines and Implementation Details}
\label{sec4_2:baseline}
\paragraph{Baselines.}
For baseline models, we include widely-used GNN-based methods: GCN~\cite{kipf2016semi}, GAT~\cite{velivckovic2017graph}, and ChebyNet~\cite{defferrard2016convolutional}.
We also include a 2-layer MLP as a baseline since MLP models show comparable performance under heterophily when existing methods for graph-structured data do not use the graph topology effectively.
To compare our models with state-of-the-art models on heterophilic graphs, we include JKNet~\cite{xu2018representation}, MixHop~\cite{abu2019mixhop}, $\text{H}_2$GCN~\cite{zhu2020beyond}, and Geom-GCN~\cite{pei2020geom}.
We use the best models among the three variants of Geom-GCN from the paper~\cite{pei2020geom}. 

\paragraph{Implementation details.}
We use the Adam optimizer~\cite{kingma2014adam} with $\ell_2$-regularization and 500 epochs for training our model and the baselines.
For all cases, hyperparameters are optimized in the same search space: learning rate in \{0.01, 0.005\}, weight decay in \{1e-3, 5e-4, 5e-5\}, and hidden dimension in \{32, 64\}.
Dropout is applied with $0.5$ dropout rate.
The performance is reported with the best model on the validation datasets.
For all datasets, we apply the splits~(48\%/ 32\%/ 20\%)\footnote[1]{https://github.com/graphdml-uiuc-jlu/geom-gcn.} of nodes per class for (train/ validation/ test) provided by \cite{pei2020geom} for a fair comparison as \cite{zhu2020beyond}.
All experiments are repeated 10 times as \cite{pei2020geom, zhu2020beyond} and accuracy is used as an evaluation metric.
More implementation details are in the supplementary materials.
\subsection{Results on Node Classification}
Table~\ref{tab:main table} shows the results of Deformable GCN and other baselines on node classification tasks.
The best model for each dataset is highlighted with boldface.
Overall, Deformable GCN achieves the highest performance on all heterophilic graphs compared to all baselines including $\text{H}_2$GCN, which is specifically proposed for heterophilic graphs.
Note that on some heterophilic graph datasets such as Texas, Wisconsin, Actor, and Cornell, MLP outperforms various GNNs such as GCN, GAT, and Geom-GCN by significant margins without utilizing any graph structure information.
It might seem that graph structure information is harmful to representation learning on heterophilic graphs for most existing GNN models, but on other heterophilic graph datasets such as Squirrel and Chameleon, MLP underperforms most GNN models. 
In contrast to the existing GNN models, 
our Deformable Graph Convolution in various \textit{position spaces} with \textit{node positional embeddings} allows Deformable GCN to consistently achieve the best performance on all the heterophilic graph datasets. 

Similar to our method, Geom-GCN also performs convolution in a latent space. But it significantly underperforms our method by 17.4\% on average on heterophilic graph datasets and especially on Cornell the gap is 25.4\%.
We believe that our node positional embeddings, which are simultaneously learned with node representations for the target task in an end-to-end fashion, are more effective than the node embeddings in Geom-GCN that are obtained from pre-trained external embedding methods.
On homophilic graphs, Deformable GCN shows comparable performance.
Since most existing GNNs have been proposed based on the homophily assumption, the performance gap between GNN-based models is small.



\subsection{Ablation Study and Analysis}
\begin{figure}[t!]
\centering

\includegraphics[width=1.0\linewidth]{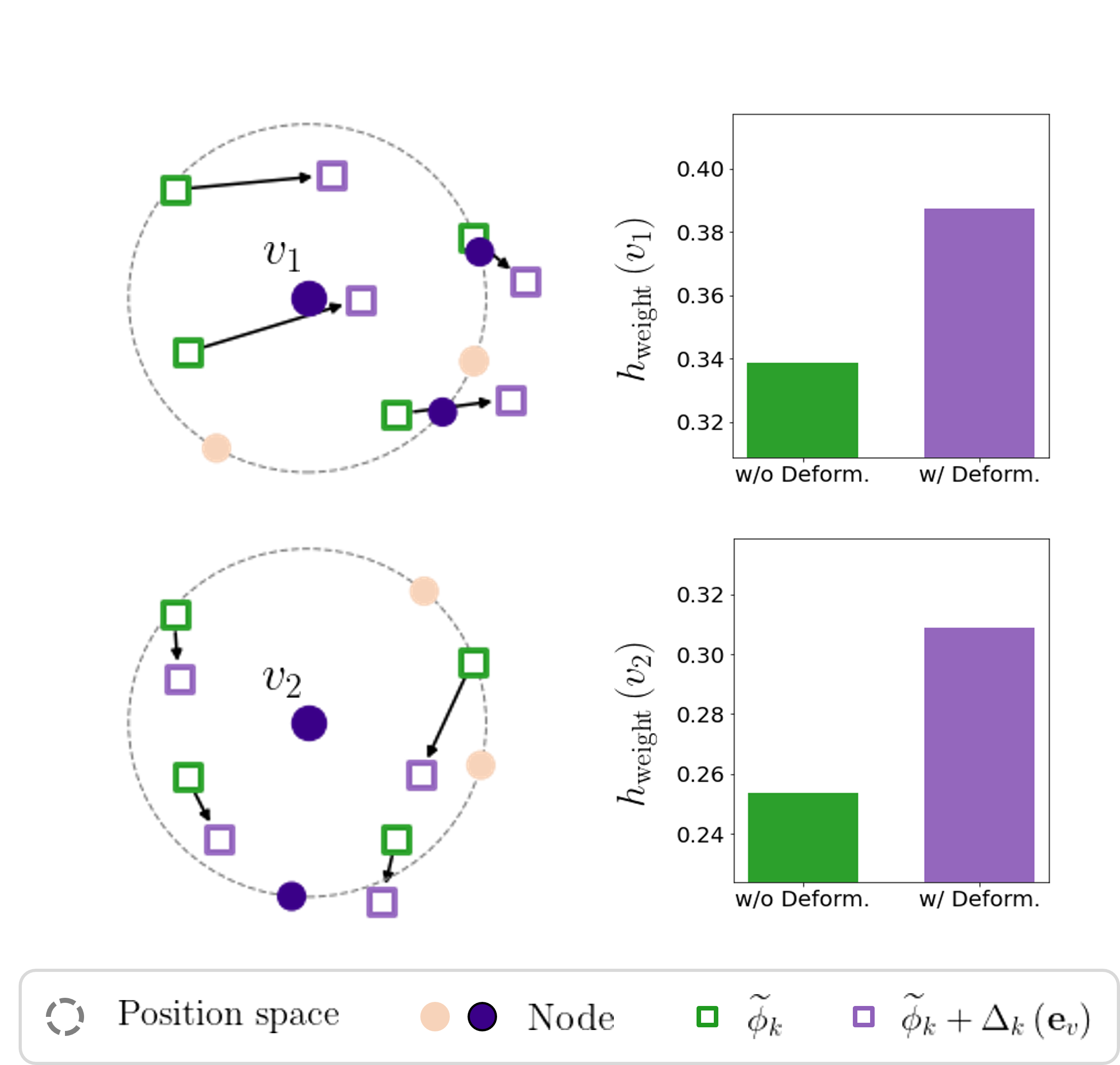}

\caption{(Left) The visualization of kernel vectors without deformation~$\widetilde{\phib}_k$ and kernel vectors with deformation ~$\widetilde{\phib}_k + \Delta_k\left(\eb_v\right)$ in a position space centered at each node $v_1$ and $v_2$.
The color of a node indicates its label. 
The kernel vectors are differentially deformed to increase the homophilic weight $h_{\text{weight}}\left(v \right)$ approaching neighbor nodes with the same labels (black solid circles).
(Right) Comparisons of the homophilic weight $h_{\text{weight}}\left(v\right)$ defined in \eqref{eq_sup:weight} between w/o and w/ Deformation. 
}
\label{fig_sup:deform}
\end{figure}

\jyp{We conduct additional experiments to verify the contributions of our node positional embedding, deformable graph convolution layer, deformation, and regularization for Deformable GCN and analyze the attention score $\left\{s^{(l)}\right\}_l$ and the receptive field of Deformable GCN.

\begin{table}[t]
    \centering
    \setlength{\tabcolsep}{4pt}
    \begin{tabular}{c|cccc}
    \toprule
    Positional&\multicolumn{4}{c}{Dataset}\\
    embedding& Wisconsin & Actor & Squirrel & Pubmed  \\
    \midrule
    node2vec        &   67.57 & 35.04  &    45.27 & 88.35  \\
    PoinCare        & 68.1        & 35.15 & 46.31 & 87.26    \\
    Ours            &\textbf{87.06} & \textbf{37.07}  &    \textbf{62.56} & \textbf{89.49}  \\
    \bottomrule
  \end{tabular}
  \caption{Comparison of our node positional embeddings with other node embedding methods on four datasets.}
  \label{tab:pos_abl}
\end{table}

\begin{table}[t]
    \centering
    \small
    \setlength{\tabcolsep}{4pt}
    \begin{tabular}{c|cccc}
    \toprule
    &\multicolumn{4}{c}{Dataset}\\
    Layer& Wisconsin & Actor & Squirrel & Pubmed \\
    \midrule
    GAT Layer        & 70.54 & 36.26  &   61.62 & 88.04 \\
    Deformable GConv        & \textbf{87.06} & \textbf{37.07} & \textbf{62.56} & \textbf{89.49} \\
    \bottomrule
  \end{tabular}
  \caption{\label{tab_sup:gatlayer_abl}Comparison of our Deformable GConv with GAT
Layer on four datasets.}
\end{table}
\begin{figure}[!t]
\centering
\subfigure[Wisconsin]{
\includegraphics[width=0.47\linewidth]{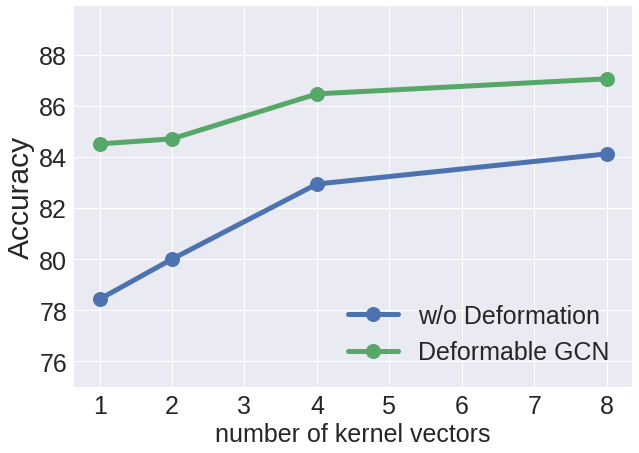}
}
\subfigure[Pubmed]{
\includegraphics[width=0.47\linewidth]{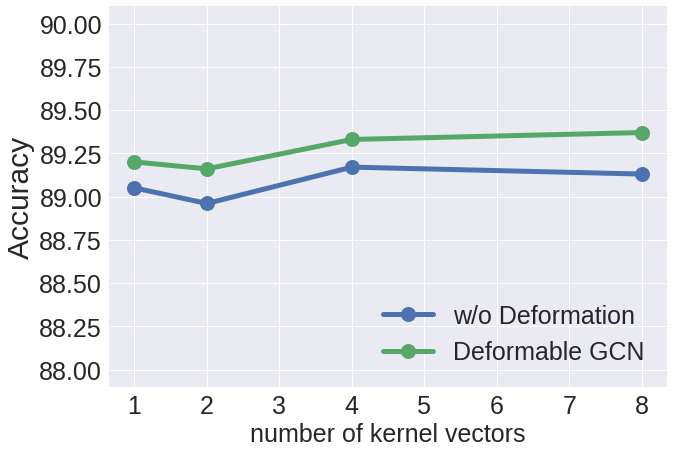}
}
\caption{Ablations for deformation of deformable graph convolution.
We compare \textbf{Deformable GCN} and the model without deformation~(\textbf{w/o Deformation}), which uses only kernel vectors on (a) Wisconsin and (b) Pubmed datasets according to number of the kernel vectors.
}
\label{fig:abl}
\end{figure}

\paragraph{Node positional embedding.}{
To verify the efficacy of our node positional embedding method, we compare it with other node embedding methods such as node2vec \cite{grover2016node2vec} and Poincare embedding \cite{nickel2017poincare} by using them for $\phib_v$ in \eqref{eq:position} on four datasets in Table~\ref{tab:pos_abl}.
The result shows that our proposed embedding method outperforms two other embedding methods for each dataset.
Specifically, our proposed embedding method improves over node2vec and Poincare embedding by 9.9\% and 9.8\% on average, respectively.
Unlike node2vec and Poincare embeddings that require separate pre-training, our positional embedding can be simultaneously trained with GNNs in an end-to-end fashion.
We believe that our embedding scheme is much easier to train and allows more optimized positional embeddings for target tasks.}

\paragraph{Deformable graph convolution.}{
In Table~\ref{tab_sup:gatlayer_abl}, we conduct an experiment to examine the contribution of deformable graph convolution~(Deformable GConv) by substituting it with the GAT Layer in Graph Attention Networks~\cite{velivckovic2017graph} on four datasets. 
From the table, Deformable GConv consistently shows better performance compared to GAT Layer.
In particular, on Wisconsin dataset, Deformable GConv improves 16.52\% over GAT Layer.
It means that Deformable GConv contributes to the performance improvements of Deformable GCN.
}

\paragraph{Deformation.}
{To reveal the effectiveness of deformation, we compare the results of Deformable GCN and the model without deformation~(w/o Deformation) on node classification task with Wisconsin and Pubmed datasets according to the number of kernel vectors. 
Figure~\ref{fig:abl} shows that Deformable GCNs consistently achieve  superior performance~($2\sim6 \%$ on average) than the models without deformation~(w/o Deformation) on both a heterophilic graph Wisconsin and a homophilic graph Pubmed in the various settings with different numbers of kernel points.

Also, for qualitatively analyzing the effectiveness of deformation, we visualize the kernel vectors and deformation in the position space in Figure \ref{fig_sup:deform}.
To study the efficacy of the deformation, we calculate the homophilic weight based on $a_{u,v,k}$ in \eqref{eq:ours_conv} as follows:
\begin{equation}
\label{eq_sup:weight}
    h_{\text{weight}}\left(v\right) = \sum_{\left\{u : u \in N\left(v\right) \land y_u = y_v\right\}}\sum_k a_{u,v,k}.
\end{equation}
It indicates the summation of the weight $a_{u,v,k}$ of neighbor nodes with the same label as center node $v$.
In Figure \ref{fig_sup:deform}, we compare $h_{\text{weight}}\left(v\right)$ before/after deformation.
After deformation, the kernel vectors tend to become closer to the neighbor nodes with the same labels, which are black solid circles.
As a result $h_{\text{weight}}\left(v \right)$ becomes larger values and Deformable GCN receives more messages from meaningful neighborhoods (e.g., with the same labels).
}}

\begin{table}[t]
    \centering
    \setlength{\tabcolsep}{4pt}
    \begin{tabular}{cc|cccc}
    \toprule
    \multicolumn{2}{c|}{Regularizer} & \multicolumn{4}{c}{Dataset}\\
    $\mathcal{L}_{sep.}$ & $\mathcal{L}_{focus}$    & Wisconsin & Actor & Squirrel & Pubmed  \\
    \midrule
       &    & 84.12        & 36.67 & 60.31 & 89.02    \\
    \checkmark  &   & 86.08 & 36.70  &    60.43 &  88.86 \\
      &\checkmark   & 86.08        & 36.67 & 61.83 & 88.79    \\
    \checkmark  &\checkmark    &\textbf{87.06} & \textbf{37.07}  &    \textbf{62.56} & \textbf{89.49}  \\
    \bottomrule
  \end{tabular}
  \caption{Ablations for two regularizer losses ($\Lc_{sep.}, \Lc_{focus}$) on four datasets.}
  \label{tab:reg_abl}
\end{table}
\begin{figure}[t]
\centering

\includegraphics[width=1.0\linewidth]{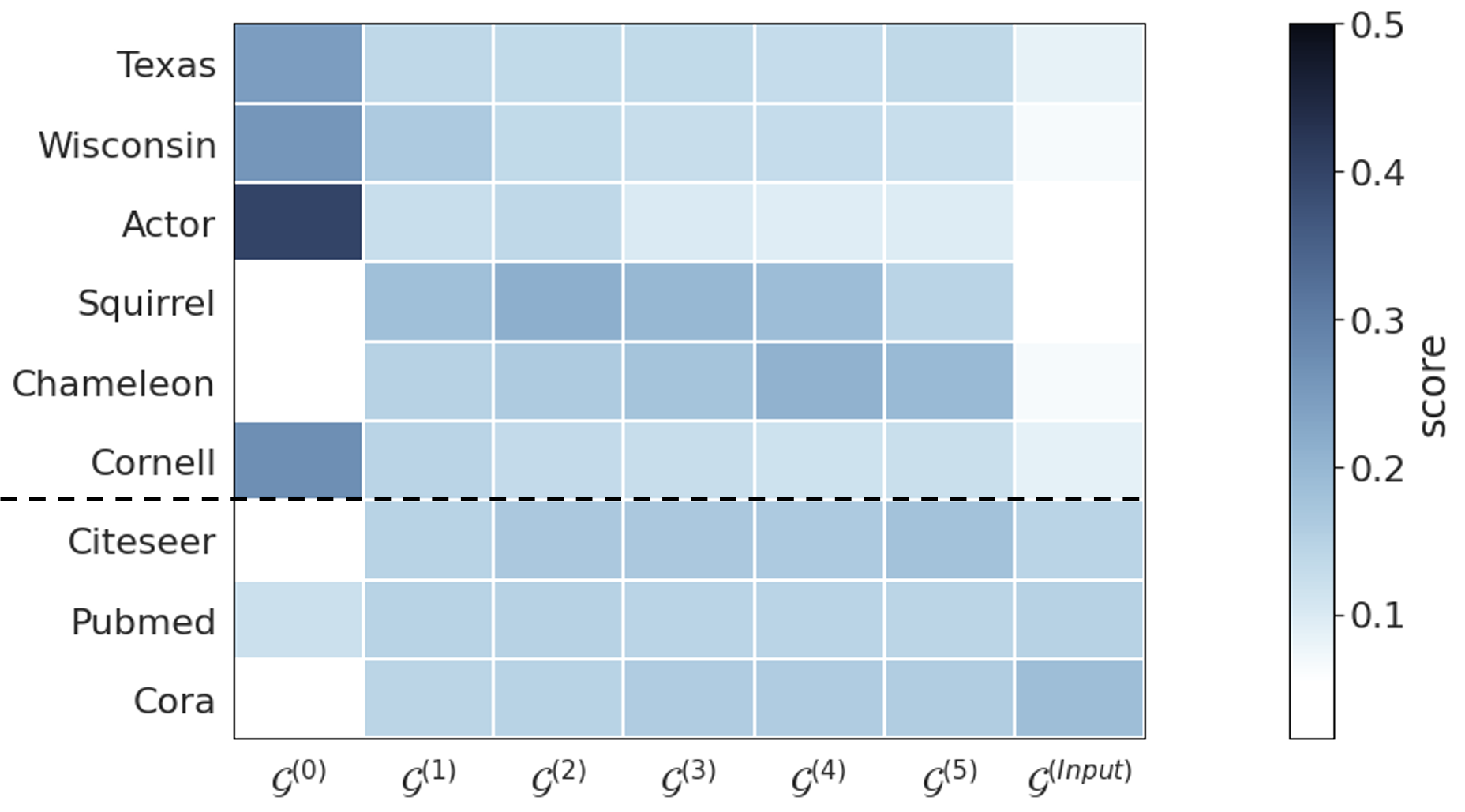}

\caption{The attention score $s^{(l)}$ of each \textit{latent neighborhood graph} $\Gc^{(l)}$. 
Datasets above the dash line are heterophilic graphs and the others homophilic graph datasets.
}
\label{fig:atten}
\end{figure}

\paragraph{Regularizers.}
To verify contribution of regularizers, we conduct an ablation study on regularizers such as $\Lc_{sep.}$ and $\Lc_{focus}$ on four datasets.
Table~\ref{tab:reg_abl} summarizes the results of the ablation study of our regularizers.
From the table, each regularizer contributes to improving performance of Deformable GCN.  
Moreover, we observe that training with both regularizers is the most effective.
On average, Deformable GCNs trained with both regularizers outperform ones without regularizers by 1.5\%.

\begin{figure}[t]
\centering

\includegraphics[width=1.0\linewidth]{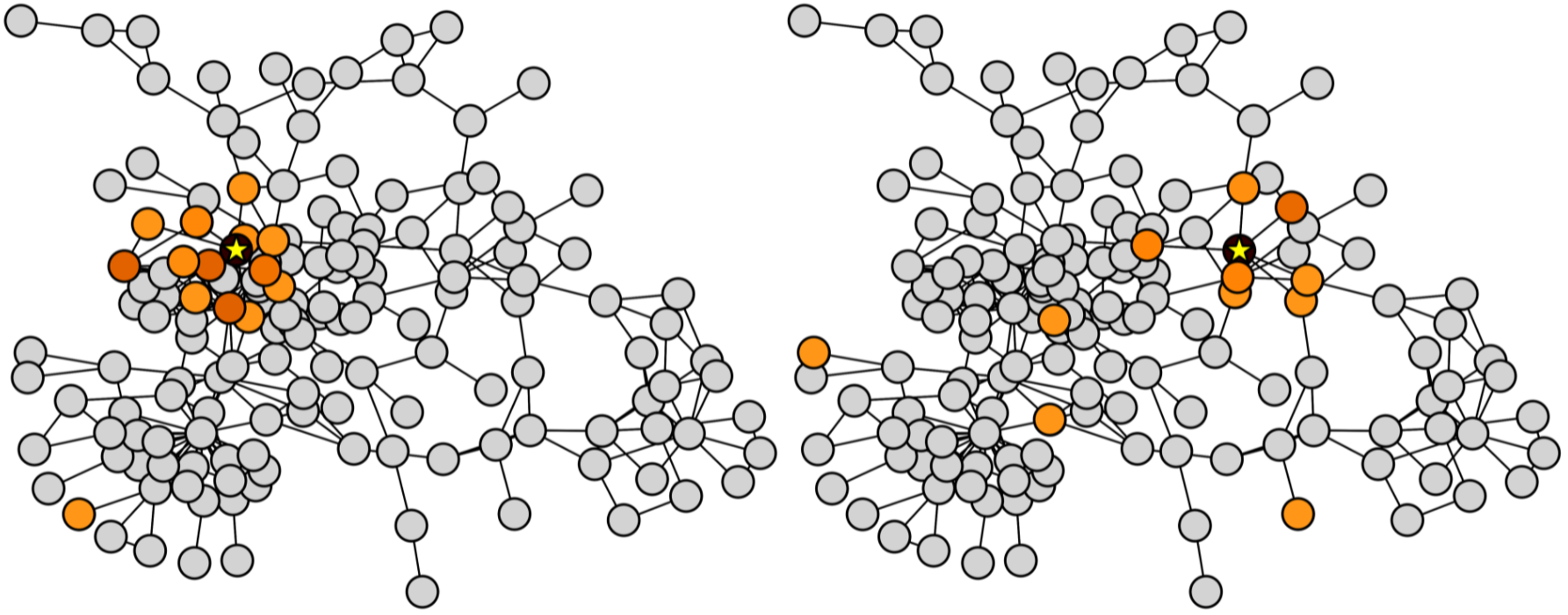}

\caption{The visualization of the receptive field of each target node marked with star on Cora dataset. 
The intensity of the color of each node is associated with the intensity of each node. 
}
\label{fig:recep}
\end{figure}

\paragraph{Attention score $s_v^{(l)}$.}{
The attention score $s_v^{(l)}$ in \eqref{eq:score} at node $v$ for a latent neighborhood graph $\Gc^{(l)}$ can be used to
understand datasets. 
An averaged attention score $s^{(l)} = \frac{1}{|V|} \sum_{v \in V} s_v^{(l)}$ indicates the overall importance of a latent neighborhood graph $\Gc^{(l)}$.
Figure~\ref{fig:atten} shows the attention score of a Deformable GCN with $L=5$ that utilizes 6 latent neighborhood graphs $\{\Gc^{(l)} \}_{l=0}^{l=5}$ and a original input graph $\Gc^{(\text{Input})}$.
On most heterophilic graph datasets except for Chameleon and Squirrel, 
Deformable GCN has the large attention score for latent graph $\Gc^{(0)}$ and relatively small attention scores for $\Gc^{(\text{Input})}$.
Recall that $\Gc^{(0)}$ is the kNN graph constructed based on the similarity between input features.
This indicates that rather than focusing on the neighborhood on the input graph, smoothing over the nodes with similar input features is more helpful for predictions on heterophilic graphs. 
Indeed, this coincides with the results in Table \ref{tab:main table}. On the heterophilic graphs such as Texas, Wisconsin, Actor, and Cornell, an MLP, which does {\em not} use any graph structure information, outperforms most existing GNNs.
On the other hand on homophilic graphs, $\Gc^{(\text{Input})}$ plays a more important role on homophilic graphs compared to heterophilic graphs.
}

\paragraph{Receptive field of Deformable GCN.}{
We visualize the receptive field for each target node, which is marked as yellow star, on the subgraph of Cora dataset in Figure~\ref{fig:recep}.
For visualization, we compute the intensity of node $u$ as $\sum_l \sum_k s_{v}^{(l)}\cdot \widehat{a}_{u,v,k}$, where the node $v$ is target node.
We represent the intensity of the color of each node according to the intensity of each node.
The figure shows that the receptive field of Deformable GCN varies depending on the target node on a single graph, which means that our model flexibly captures various ranges of node depenedencies as needed.
}



\section{Conclusion}
We proposed Deformable Graph Convolutional Networks for learning representations on both heterophilic and homophilic graphs.
Our approach learns node positional embeddings for mapping nodes into latent spaces and applies deformable graph convolution on each node.
The convolution kernels in deformable graph convolution are transformed by shifting kernel vectors.
Our experiments show that the Deformable Graph Convolution Networks are effective for learning representations on both heterophilic and homophilic graphs.
Future directions include studying the effectiveness of Deformable GCN in other tasks such as link prediction and graph classification and expanding Deformable GConv to other domains.

\paragraph{Limitations.}
In our work, for all datasets, we generate the node positional embeddings with a simple procedure that is smoothing input features followed by a linear transformation.
We believe that for the datasets with more complicated input features, the simple procedure may not be sufficient and more advanced techniques might need to be developed.
\section*{Acknowledgments}
This work was partly supported by MSIT (Ministry of Science and ICT), Korea, under the ICT Creative Consilience program (IITP-2021-2020-0-01819) supervised by the IITP and Samsung Research Funding \& Incubation Center of Samsung Electronics under Project Number SRFC-IT1701-51.

{
\bibliography{reference}
}

\end{document}